%% file: eacl2023.tex
\pdfoutput=1

\documentclass[11pt]{article}

\usepackage{EACL2023}

\usepackage{times}
\usepackage{latexsym}

\usepackage[T1]{fontenc}

\usepackage[utf8]{inputenc}

\usepackage{microtype}

\usepackage{inconsolata}

\usepackage{amsmath}
\usepackage{booktabs}
\usepackage{lscape}
\usepackage{multirow}
\usepackage{array,graphicx}
\usepackage{flexisym}
\newcommand{\STAB}[1]{\begin{tabular}{@{}c@{}}#1\end{tabular}}

\newcommand{\dist}{\mathrm{d}}

\newcommand\blfootnote[1]{%
  \begingroup
  \renewcommand\thefootnote{}\footnote{#1}%
  \addtocounter{footnote}{-1}%
  \endgroup
}

\title{Neural Machine Translation Models Can Learn to be Few-shot Learners}

\author{Raphael Reinauer$^\dagger$ \and Patrick Simianer$^\dagger$ \and Kaden Uhlig \and Johannes E.\ M.\ Mosig\\ \and {\bf Joern Wuebker}\\
Lilt\\
\texttt{\{raphael.reinauer,patrick\}@lilt.com}}

\begin{document}
\maketitle
\begin{abstract}
The emergent ability of Large Language Models to use a small number of examples to learn to perform in novel domains and tasks, also called in-context learning (ICL).
In this work, we show that a much smaller model can be trained to perform ICL by fine-tuning towards a specialized training objective, exemplified on the task of domain adaptation for neural machine translation. 
With this capacity for ICL, the model can take advantage of relevant few-shot examples to adapt its output towards the domain. 
We compare the quality of this domain adaptation to traditional supervised techniques 
and ICL with a 40B-parameter Large Language Model.
Our approach allows efficient batch inference on a mix of domains %
and outperforms state-of-the-art baselines in terms of both translation quality and immediate adaptation rate, i.e.\ the ability to reproduce a specific term after being shown a single example.
\end{abstract}

\section{Introduction}

\blfootnote{$^\dagger$ Equal contribution.}

Large Language Models (LLMs) have demonstrated few-shot learning capabilities on various natural language processing tasks, as highlighted by \citet{brown2020language} or \citet{garcia2023unreasonable}. 
When prompted with suitable example translations, they can compete with neural machine translation (NMT) models, built and trained specifically for translating between languages \cite{vilar2022prompting}. 
Interestingly, one can adapt LLMs to specific domains merely by adding example translations to their prompt at inference time \cite{moslem2023adaptive}.
This ability to adapt to specific tasks and domains is known as \textit{in-context learning} (ICL). 
In contrast to traditional fine-tuning methods, ICL does not require a separate set of customized parameters for each domain, which implies major efficiency gains through batched inference.

In this paper, we integrate ICL for domain adaptation into NMT systems in multiple steps. 
We compare our method for adapting NMT systems to traditional fine-tuning approaches, as well as to the domain adaptation abilities of an open-source LLM.
Specifically, our main contributions are the following:
\begin{enumerate}
    \item We evaluate an unmodified NMT system's ICL capacity for domain adaptation and demonstrate its limitations.
    \item We propose a training scheme to improve an NMT model's ICL capability.
    \item We show that ICL can be combined with more traditional adaptation methods to further improve domain adaptation performance.
    \item We compare our method to the performance of the open-source LLM \textsc{Falcon}-40B \cite{penedo2023refinedweb} on a machine translation task with ICL for domain adaptation.
\end{enumerate}

\section{Related Work}

\citet{bulte-tezcan-2019-neural} improve the translation performance of an NMT model by integrating translation fuzzy-matched pairs from a translation memory as input to an NMT model. 
This idea was further expanded by \citet{Pham2020PrimingNM} and \citet{xu-etal-2020-boosting}, who for a given source segment use sentence embeddings to retrieve similar examples and compared different schemes for integrating those as cues into the NMT network. 

Our approach differs in that we only train on the tokens belonging to the translation and not on the tokens provided as context, which we show to work better.
In addition, \citet{Pham2020PrimingNM}'s training procedure differs, as they train their model from scratch, using training data from multiple domains and evaluate on those same domains, while we train on general domain data and evaluate on a new domain that is not in the training data. 
Furthermore, we focus on the multi-domain adaptation task using light-weight adapters. 
This approach not only allows us to extend to new domains without retraining the full model, but also offers a more practical and efficient strategy for real-world applications.

The authors of \cite{moslem2023adaptive} investigated the capabilities of a proprietary LLM, specifically GPT-3.5, for adaptive machine translation using ICL. 
Their extensive experiments showed that GPT-3.5 can adapt well to in-domain sentence pairs and/or terminology.

\section{Experiments}

We conduct a series of experiments to develop NMT systems that exceed at few-shot ICL domain adaptation. 
Here we present the experiments in a logical order, where we start with the baseline models described in Section~\ref{sec:exp:models} and subsequently introduce several stages of development. 
In stages 0 and 1 we attempt ICL with the unmodified and domain-fine-tuned baseline models, respectively. 
Then, in \textsc{Stage~2}, we fine-tune the baseline model to the \textit{task} of domain ICL, instead of a particular domain.
Finally, we combine ICL and domain adaptation through fine-tuning in \textsc{Stage~3}. 
Our experimental progression was guided by the metrics and datasets that we introduce in Sections \ref{subsec:exp:metrics} and \ref{subsec:exp:datasets}, respectively. 

\subsection{Models}
\label{sec:exp:models}

Throughout this paper, we work with an NMT system and the \textsc{Falcon-40B} LLM, which we both describe here.

\subsubsection{\textsc{Falcon} LLM}

To provide a direct comparison with LLMs and their capacity for ICL, we conduct experiments with the decoder-only Transformer language model \textsc{Falcon-40B} \cite{penedo2023refinedweb}, specifically the non-instruction-tuned variant\footnote{The model  is available from the \textit{huggingface} platform: \url{https://huggingface.co/tiiuae/falcon-40b}}. Inference is done with greedy decoding. 
Following previous work \cite{bawden2023investigating, garcia2023unreasonable, hendy2023good} (\textit{inter-alia}) the model is prompted to perform translation without specific fine-tuning towards the machine translation task.

A simple prompt template is used for all $k$-shot experiments with \textsc{Falcon-40B}, see Figure \ref{fig:prompt}.

\begin{figure}
\begin{verbatim}
    English: <source sentence>\n
    German: <target sentence>\n
    English: [...]
\end{verbatim}
\caption{Prompt template for LLM.}
\label{fig:prompt}
\end{figure}

In preliminary experiments we found that $k=0$ does not work well with this specific model\footnote{For $k=0$ the prompt contains only the single source sentence as input and the target language followed by a colon.} -- the outputs tend to be entirely hallucinated.

\subsubsection{NMT Systems}

The baseline model that we use as the starting point for all further experiments is a Transformer \cite{vaswani2017attention} model with 12 encoder layers and two decoder layers, implemented with the NVIDIA NeMo toolkit \cite{kuchaiev2019nemo}. 
The embedding size is 1,024 with a feed-forward network dimension of 4,096. 
The model has a joint vocabulary of 32,768 tokens, while embedding matrices are specific to the encoder, decoder, and output projection modules, i.e.\ parameters are not shared between them. 
The model was trained to support a maximum input size of 1,536 tokens by augmenting the training data with random concatenations of parallel sentences. 
We evaluate the model using greedy decoding.

For the experiments presented here, the baseline model is either fine-tuned in full (\textsc{Stage~2a} and \textsc{Stage~2b}), or light-weight adapters \cite{bapna-firat-2019-simple} are added to the model (\textsc{Stage~1} and \textsc{Stage~3}). 
We choose full-model fine-tuning on out-of-domain data to adapt the NMT model to a new task -- translating with an increased context of related examples -- and adapter layers for learning from in-domain data. 

The adapters we use follow \citet{adapter}'s formulation, but with layer normalization applied after the bottleneck rather than before it. We use a bottleneck width of 256 and insert adapters in every layer of the decoder and every other layer of the encoder.

We always fine-tune with the ADAM optimizer \cite{kingma2017adam} and early stopping based on validation loss.

\subsection{\textsc{Stage~0} \& \textsc{Stage~1}: ICL with a Standard NMT Model}\label{subsec:exp:icl_with_std_nmt}

Motivated by the few-shot learning capabilities of LLMs, we examine the ability of a standard English-to-German NMT model to adapt to a domain given only similar and relevant translation pairs as additional context, i.e., without changing the model's parameters.

To find similar source segments in the translation memory, we search for nearest neighbours in an embedding space. 
We use the multi-lingual sentence embedding model\footnote{Model name on \url{https://www.sbert.net/}: \texttt{all-MiniLM-L6-v2}} from the sentence transformer library \cite{reimers-2020-multilingual-sentence-bert} to embed the source sides of all segment pairs. 
Then we employ \texttt{hnswlib} \cite{malkov2018efficient} to find the approximate nearest neighbours: Each source sentence in the domain-specific datasets is first encoded with the sentence-embedding model and then added to an index. For the sake of simplicity in this paper, we will refer to the approximate nearest neighbors simply as nearest neighbors.
To measure the similarity between a pair of segments \texttt{s} and \texttt{s\textprime}, we use the cosine distance of the corresponding embedding vectors $\texttt{v}_{\texttt{s}}$ and $\texttt{v}_{\texttt{s\textprime}}$, i.e., 
$$\dist(\texttt{s}, \texttt{s\textprime}):= 1- \frac{\texttt{v}_s \cdot \texttt{v}_{\texttt{s\textprime}}}{\|{\texttt{v}_{\texttt{s}}}\|_2 \cdot \|\texttt{v}_{\texttt{s\textprime}}\|_2}.$$

For a given source \texttt{s} and target segment \texttt{t}, we identify its nearest neighbours \texttt{$s_1$}, \texttt{$s_2$}, ..., \texttt{$s_k$}, using the the cosine distance above. Each source sentence \texttt{$s_i$} is paired with a reference translation \texttt{$t_i$} for $i=1,...,k$.
We sort the pairs by their distance to \texttt{s} in the embedding space, i.e., 
$$\dist(\texttt{s}, \texttt{s}_1) \leq \dist(\texttt{s}, \texttt{s}_2) \leq ... \leq \dist(\texttt{s}, \texttt{s}_k)\;.$$

Our assumption is that similar segments should have similar translations. 
For \textsc{Stage~0} of the experiments, we treat the context sentences and actual source text as one body of text, separated only by a single space, ordering the segments from least similar to most similar, with the current source segment $\texttt{s}$ at the end. As a result, the input of the encoder is 
$$
\texttt{<bos> s}_k \texttt{ s}_{k-1} \ ...\  \texttt{s}_{1} \texttt{ s <eos>}
$$
while for the decoder, we use the prefix:
$$
\texttt{<bos> t}_k \texttt{ t}_{k-1} \ ...\  \texttt{ t}_1
$$
where \texttt{<bos>} and \texttt{<eos>} represent the beginning-of-sentence and end-of-sentence tokens, respectively. The model's task is then to continue from the target prefix by generating a translation of the source segment $\texttt{s}$.

In our experiments, we evaluated the translation performance using a varying number $k$ of nearest neighbors, specifically $k \in \{1, 2, 5\}$.

In \textsc{Stage~1} we run additional experiments where we fine-tune the model for each domain, using the in-domain training data in the original format. 
This domain-specific fine-tuning is performed by injecting adapter layers \cite{bapna-firat-2019-simple} into the network while freezing the rest of the model, and leveraging a standard negative log-likelihood (NLL) loss for training.
For each domain, we then test the fine-tuned model directly  ($0$-shot in Tables \ref{tab:ACED-results} and \ref{tab:MDNS-results}) as well as with ICL ($k$-shot with $k\neq0$).

Adapters are trained towards convergence, i.e.\ until there is no further improvement in terms of validation loss.

\subsection{\textsc{Stage~2a} \& \textsc{Stage~2b}: Fine-Tuning towards ICL}\label{subsec:stage_2_exp}

To improve the model's capability to use nearest neighbor examples in the context, we further fine-tune the full model on out-of-domain data, namely \textit{News-Commentary}\footnote{From the WMT'23 evaluation campaign: \url{https://data.statmt.org/news-commentary/v18.1/}}  \citep{kocmi-etal-2022-findings}, which contains roughly 450K parallel segments. For validation we use a sample of 2K parallel segments from \textit{EuroParl}\footnote{Also from the WMT'23 evaluation campaign: \url{https://www.statmt.org/europarl/v10/}} \citep{koehn2005europarl}.
For this full model fine-tuning we do not train until convergence, but apply aggressive early stopping: Training is stopped when the validation loss does not decrease by at least 0.1 twice in a row, validating for every 1\% of an epoch. This is to encourage the model to only learn the new task and data format, but not adapt to a new data distribution.

Instead of directly concatenating the nearest neighbors to the training examples, we add a special separation token -- \texttt{<sep>} -- to separate the source and target segments. 
We then construct the training instances for the encoder as:
$$
\small
\texttt{<bos> s}_k \texttt{ <sep> s}_{k-1} \ \texttt{<sep>} \ ...\ \texttt{<sep> s}_{1} \texttt{ <sep> s <eos>}
$$
and for the decoder as:
\begin{align}\label{eq:few_shot_target_with_sep}
\small
\texttt{<bos> t}_k \texttt{ <sep> t}_{k-1} \ \texttt{<sep>} \ ...\  \texttt{<sep> t}_{1} \texttt{ <sep> t <eos>}
\end{align}
and compute the NLL loss on all tokens of (\ref{eq:few_shot_target_with_sep}). 
This training loss is identical to the one used in \citet{Pham2020PrimingNM}. We denote this procedure as \textsc{Stage~2a}.

For \textsc{Stage~2b} the idea is that the model should learn to predict the target segment from the source segment using the nearest neighbor translations but not learn to predict \texttt{$t_k, ..., t_1$} as in \cite{Pham2020PrimingNM}. 
Hence we mask the NLL training loss such that it is computed only on the tokens that belong to the target segment \texttt{t}, excluding all context tokens, thus fully focusing the training signal on translating \texttt{t} in the context of its $k$ nearest neighbors.

We then use the same format as in \textsc{Stage~2a} for training, while at inference time we provide the decoder with a prefix containing the ICL examples:
$$
\small
\texttt{<bos> t}_k \texttt{ <sep> t}_{k-1} \ \texttt{<sep>} \ ...\  \texttt{<sep> t}_{1} \texttt{ <sep>}
$$
Finally, we measure quality of the predicted translation $\hat{t}$ by computing BLEU and COMET scores with the target segment \texttt{t} as reference.

For both \textsc{Stage 2a} and \textsc{Stage 2b}, the $k$-nearest neighbors for each segment in the training data and validation data are extracted from the entire \textit{News-Commentary} dataset as described in Section \ref{subsec:exp:icl_with_std_nmt}.

\subsection{\textsc{Stage~3}: Combining ICL and Domain Adaptation}

To combine \textsc{Stage~2b}'s ICL capacity with adapter-based domain adaptation, we add adapters to the model from \textsc{Stage~2b} using the same configuration as for the \textsc{Stage~1} experiments. 
Again, we train separate adapter layers for each domain.

Each example from the training set is annotated with its nearest neighbors from the same training set, excluding itself.

\subsection{Metrics}
\label{subsec:exp:metrics}

For evaluating translation quality, we used the SacreBLEU framework \cite{post-2018-call} that implements the BLEU metric \cite{papineni-etal-2002-bleu-1}. We also evaluate with reference-based COMET \cite{rei-etal-2022-comet} to compare the model outputs to the reference translations in the test data.

\subsection{Datasets}
\label{subsec:exp:datasets}

We run our experiments with the English-German language pair on 8 domains from the ACED- and MDNS corpus collections, which we describe in this section.
Statistics for all datasets are provided in \autoref{tab:dataset_stats}.

\subsubsection{ACED corpus}

The ACED corpus \cite{lin-etal-2022-automatic} is comprised of three distinct datasets, namely Asics, Emerson, and Digitalocean, each consisting of English-German sentences extracted from various domains. 
ACED is a real-world benchmark containing data derived from translations performed by humans. 

\subsubsection{MDNS corpus}

The MDNS corpus \cite{aharoni2020unsupervised} is a multi-domain corpus containing English-German parallel text from five diverse domains (IT, Koran, Law, Medical, Subtitles). 
It was specifically created for evaluating domain-adaptation. 

\begin{table}
\centering
\begin{tabular}{c|r|r|r}
& Training & Validation& Test\\ 
\hline
Asics & 1.4 & 0.5 & 0.6 \\
Digitalocean & 11.8 & 2.0 & 7.6 \\
Emerson & 4.3 & 1.3 & 1.7 \\
\hline
\hline
IT & 223 & 2.0 & 2.0\\
Koran & 17.9 & 2.0 & 2.0 \\
Law & 467 & 2.0 & 2.0 \\
Medical & 248 & 2.0 & 2.0 \\
Subtitles & 500 & 2.0 & 2.0 \\
\end{tabular}
\caption{Segment counts for the domain-specific dataset splits used for experimentation, in thousands.}
\label{tab:dataset_stats}
\end{table}

\begin{table*}
\centering
\begin{tabular}{c c|c|c||c|c|c|c|c}
& \multicolumn{3}{c}{ACED} & \multicolumn{5}{c}{MDNS}\\
& Asics & Digitalocean & Emerson & IT & Koran & Law & Medical & Subtitles \\
\hline
$k=1$ & 0.19 & 0.30 &  0.13 & 0.15 & 0.18 & 0.13 & 0.12 & 0.24 \\
$k=2$ & 0.21 & 0.31 & 0.14 & 0.17 & 0.20 & 0.15 & 0.14 & 0.25 \\
$k=5$ & 0.23 & 0.34 & 0.16 & 0.21 & 0.24 & 0.17 & 0.17 & 0.27 \\
\end{tabular}
\caption{Average cosine distance in embedding space of test set sources to $k$-nearest neighbors from train, for $k \in \{1,2,5\}$.}
\label{tab:dists}
\end{table*}

\section{Results}

Here we discuss the experimental results, progressing from \textsc{Stage~0} to \textsc{Stage~3}. 
All results are depicted separately for ACED-\ and MDNS corpora in Tables~\ref{tab:ACED-results} and \ref{tab:MDNS-results} respectively. 

\begin{center}
\def\arraystretch{1.4}
\begin{table*}[!ht]
\setlength{\tabcolsep}{2pt}
\footnotesize
    \centering
    \input{results_table_tec}
    \caption{Results for the ACED corpus of the multi-stage evaluation for various numbers of $k$-nearest-neighbors, using BLEU and COMET metrics. %
    The "Baseline" scores are for the English-to-German NMT system described in Section~\ref{sec:exp:models}. %
    We omit the Digitalocean dataset for the \textsc{Falcon-40B} 5-shot evaluation.}
    \label{tab:ACED-results}
\end{table*}
\end{center}

\begin{center}
\def\arraystretch{1.4}
\begin{table*}[!ht]
\setlength{\tabcolsep}{1pt}
\footnotesize
    \centering
    \input{results_table_mdns}
    \caption{Results for the MDNS corpus of the multi-stage evaluation for various numbers of $k$-nearest-neighbors using BLEU and COMET metrics. %
    The "Baseline" scores are for the English-to-German NMT system described in Section~\ref{sec:exp:models}.}
    \label{tab:MDNS-results}
\end{table*}
\end{center}

\subsection{\textsc{Stage~0}: ICL with Baseline NMT Model}

When we add nearest neighbors to the inputs and target prefixes we first observe that the automated metrics are mostly improved across all datasets. 
Notably, the result with 1-shot nearest neighbors is the best in this group of experiments.
Additionally we find that the 5-shot result often degrades below the baseline. 

Specifically for the Medical and Subtitles corpora of MDNS, we find that the model fails to improve over the baseline for all $k$. 

The cosine distance of the nearest neighbors seems to be a viable indicator of performance in this set of experiments, e.g.\ when comparing the results for ACED Emerson \& Digitalocean, where the average cosine distance (see Table~\ref{tab:dists}) for $k=1$ is much lower for Emerson at 0.13, versus 0.3 for Digitalocean. 
We find a moderate, statistically insignificant, negative Pearson correlation ($r=-0.43$) between the average cosine distances for $k=1$ and the difference in BLEU scores between the \textsc{Stage 0} 1-shot experiment and the baseline.

While BLEU indicates improvement (COMET reduces only for $k>1$), we find that the model's behavior is in fact degenerate.
Specifically, the model often fails to produce any output after the given prefix and instead predicts \texttt{<eos>} immediately, which leads to empty translations. We find that the rates of empty translations are 8.5\%, 8.1\%, and 9.1\% for $k=1, 2$, and 5 respectively. In contrast, the baseline system has a 0\% rate of empty outputs.
This is despite the model being specifically trained to support inputs covering the full context-width in pre-training.

\subsection{\textsc{Stage~1:} Combining ICL with Domain Fine-Tuning}

For \textsc{Stage~1} we first observe that the model can be effectively adapted to each domain by training adapters (see the \textsc{Stage~1}, 0-shot results in Tables~\ref{tab:ACED-results} and \ref{tab:MDNS-results}). 
A notable exception is MDNS Subtitles, where adaptation only slightly improves over the baseline. 
This result is, however, consistent with other work \cite{aharoni2020unsupervised}.

When we combine the trained adapters with ICL, we find no improvements over \textsc{Stage 1}'s 0-shot results, with the exception of ACED Asics.

Performance drops catastrophically for the MDNS Medical \& Subtitles corpora. 
The rate of empty translations also increases dramatically\footnote{Empty translation rates of \textsc{Stage~1} for each $k$ over all corpora: 1-shot: 20.0\%, 2-shot: 20.6\%, 5-shot: 13.6\%.}, with a rate of up to 63.1\% for the 1-shot result on MDNS Medical (up from 8.0\% at \textsc{Stage~0}).

\subsection{\textsc{Stage~2a} \& \textsc{Stage~2b}: Fine-Tuning towards ICL}

When we compare the \textsc{Stage~2b} (fine-tuning with the masked loss as described in Section~\ref{subsec:stage_2_exp}) to the \textsc{Stage~0} results, we find that adding the separator and fine-tuning the model leads to generally improved scores on the ACED corpora for all $k$.

BLEU Results on MDNS corpora show slightly worse performance compared to the \textsc{Stage~0} results in 3 out of 5 corpora for $k=1$, but the averages are still improved. COMET scores are however consistently improved for this comparison.
We also find that the scores for $k=2$ and $k=1$ are very close, with 2-shot being ahead of 1-shot by 0.6\% BLEU points on average on ACED data, and 1-shot being ahead of 2-shot by 0.2 BLEU points on MDNS. Which is in contrast to what we have observed in \textsc{Stage 0}.
$k=5$ still performs worse, but we observe high relative gains compared to the 5-shot \textsc{Stage~0} result.

When comparing \textsc{Stage 2a} and \textsc{Stage 2b}, i.e. the masked loss and the standard NLL loss the results are inconclusive.

We further observe that \textsc{Stage~2b} exhibits almost negligible rates of producing empty translations, at 0.3\%, 0.8\%, and 1.2\% for $k=1,2,5$ respectively.

\subsection{\textsc{Stage~3:} Combining ICL and Domain Adaptation}

When combining ICL with adapters trained with nearest neighbor annotated data, we observe the globally best results for the NMT models. 
Compared to \textsc{Stage~1}, which is also fine-tuned towards each domain, we observe greatly improved results on all automatic metrics. \textsc{Stage 3} 2-shot delivers the best result on ACED, with an improvement of 2.5 BLEU points compared to the runner-up in terms of average BLEU \textsc{Stage 1} 1-shot. On MDNS, \textsc{Stage 3} 1-shot improves over the runner-up \textsc{Stage 1} 0-shot by 3.8 points.

Especially the scores for MDNS Koran improve well above all previous models, with a relative improvement of 101\% compared to the baseline. 
The models seem to be able to make better use of close nearest neighbors in this dataset, which are often substrings of one another. See Section \ref{subsec:qa} for a detailed analysis of the copying behavior on the ACED Asics dataset.

The rate of empty translations is reduced to 0.0\% for all $k$.

We further notice that the results for 1-\ and 2-shot ICL are very similar, and that the scores for 5-shot are also improved.

\subsection{\textsc{Falcon:} Adapting Both to a Task and a Domain at the Same Time}

The \textsc{Falcon-40B} LLM proves to excel at ICL, learning a task and adapting to a domain at the same time. Notably, scores improve with higher values of $k$, which is the opposite behavior to what we have observed with NMT models. When nearest neighbors are close to the test data, as they are for the ACED Emerson and MDNS IT datasets, we find results that are close to the best \textsc{Stage 3} results. 

\textsc{Falcon-40B}'s generation speed is however very slow at an average of 2.6 tokens per second in the 1-shot setting.

Also note that we have no means at this time to check whether parts of the test data are contained in \textsc{Falcon}'s training data.

\subsection{Qualitative Analysis}
\label{subsec:qa}

Maintaining consistency in translations is an important quality criterion in the localization industry, and is a major motivator in the use of translation memories, which help ensure that marketing materials, for example, are uniform in the promised features and functions of the products being advertised \citep{emery2011multilingual}.
In NMT models, this consistency is traditionally increased by fine-tuning a translation model for a specific domain, which we denote by "\textsc{Stage~1} with 0-shot". 
In this section, we compare the fine-tuning approach with our ICL, specifically "\textsc{Stage~3} with 1-shot". 
We evaluate translation consistency on the Asics dataset.
For that purpose we select segments \texttt{s} in the test data for which the source nearest neighbor \texttt{s\textprime} in the Asics train data differs by exactly one word. 
These segments \texttt{s} are denoted as word-substitution segments. 
For each pair (\texttt{s}, \texttt{s\textprime}), we then use two sources and one target \texttt{t\textprime} in the ICL prompt and the other target \texttt{t} as reference to compare the generated translation to. 
We define the fraction of pairs for which the generated translation exactly matches the reference as the word substitution accuracy (WSA).
\begin{table*}[!ht]
\centering
\begin{tabular}{|l|p{8cm}|}
\hline
Source & Strive for every point in the women's GEL-DEDICATE ™ 6 CLAY tennis shoe by ASICS. \\
\hline
Reference Translation & Strebe nach jedem Punkt in dem GEL-DEDICATE ™ 6 CLAY Tennisschuh für Damen von ASICS. \\
\hline
\textsc{Baseline} & \colorbox{blue!30}{Mit dem} GEL-DEDICATE\colorbox{blue!30}{™} 6 CLAY \colorbox{blue!30}{Damen-Tennisschuh von ASICS kannst du jeden} \colorbox{blue!30}{Punkt erreichen.} \\
\hline
\textsc{Stage~1} with 0-shot & \colorbox{blue!30}{Mit dem ASICS} GEL-DEDICATE ™ 6 CLAY Tennisschuh für Damen \colorbox{blue!30}{kannst du jeden Punkt erreichen.}  \\
\hline
\textsc{Stage~3} with 1-shot & Strebe nach jedem Punkt in dem GEL-DEDICATE ™ 6 CLAY Tennisschuh für Damen von ASICS. \\
\hline
\end{tabular}
\caption{Comparison of example translation outputs from different models and the reference translation. %
Words that differ from the reference translation are highlighted in \colorbox{blue!30}{blue}. %
The nearest source neighbor is "Strive for every point in the men's GEL-DEDICATE ™ 6 CLAY tennis shoe by ASICS." with the reference translation "Strebe nach jedem Punkt in dem GEL-DEDICATE ™ 6 CLAY Tennisschuh für Herren von ASICS.". %
Notice that the nearest neighbor only differs by one word in each language.}
\label{tab:comparison}
\end{table*}
The results are in \autoref{tab:wsa_eval}.

\begin{table*}[!ht]
    \centering
    \begin{tabular}{|c|c|c|c|}
         \hline
         & \textsc{Stage~3} with 1-shot & \textsc{Stage~1} with 0-shot & Non-Adapted Model \\
         \hline
         Word-substitution segments & 74.60\% & 57.14\% & 1.7\% \\
         \hline
    \end{tabular}
    \caption{Results for word substitution accuracy (WSA, cf. \autoref{subsec:qa}) for various adapted and non-adapted models for word-substitution segments.}
    \label{tab:wsa_eval}
\end{table*}

The translation for \textsc{Stage~3} 1-shot achieves a WSA score of 74.6\%, compared to 57.14\% for the fine-tuning approach (\textsc{Stage~1} 0-shot), whereas the non-adapted model only produces the exact reference translation in 1.7\% of cases.

\section{Conclusions}

We have shown that a standard NMT system can be trained to be an effective in-context learner in domain adaptation tasks. 
We find that this is most effective when we combine generic fine-tuning towards the ICL task and training adapter layers for a specific domain with nearest neighbor annotated data.

When the model is not fine-tuned towards the task, we find that ICL works to some extent, but shows degenerate behavior.

While LLMs like \textsc{Falcon}-40B can adapt to the MT task with ICL, this comes at the cost of increased compute. Generally, the results with the LLM still underperform our dedicated MT models.




\bibliography{anthology,custom}
\bibliographystyle{acl_natbib}

\end{document}

%% file: results_table_tec.tex
\begin{tabular}{l|l|ll|ll|ll||ll}
 & & \multicolumn{2}{c}{Asics} & \multicolumn{2}{c}{Digitalocean} & \multicolumn{2}{c}{Emerson} & \multicolumn{2}{c}{Average} \\
 & & BLEU & COMET & BLEU & COMET & BLEU & COMET & BLEU & COMET \\
\hline
 & Baseline & 34.5 & 0.8624 & 53.3 & 0.9043 & 44.9 & 0.9108 & 44.2 & 0.8925 \\
\hline
\multirow[c]{3}{*}{\STAB{\rotatebox[origin=c]{90}{\textsc{Stage 0}}}} & 1-shot & 43.7 & 0.8578 & 54.4 & 0.8982 & 72.1 & 0.9213 & 56.7 & 0.8924 \\
 & 2-shot & 44.5 & 0.8525 & 54.5 & 0.8967 & 67.2 & 0.9137 & 55.4 & 0.8876 \\
 & 5-shot & 41.0 & 0.8420 & 53.9 & 0.8955 & 28.7 & 0.8705 & 41.2 & 0.8693 \\
\hline
\multirow[c]{4}{*}{\STAB{\rotatebox[origin=c]{90}{\textsc{Stage 1}}}} & 0-shot & 41.2 & 0.8780 & 60.1 & \textbf{0.9152} & 79.2 & 0.944 & 60.2 & 0.9124 \\
 & 1-shot & 46.4 & 0.8657 & 59.6 & 0.9099 & 78.1 & 0.9378 & 61.4 & 0.9045 \\
 & 2-shot & 46.2 & 0.8628 & 59.0 & 0.9090 & 66.3 & 0.9275 & 57.2 & 0.8998 \\
 & 5-shot & 44.2 & 0.8500 & 57.3 & 0.9038 & 32.2 & 0.893 & 44.6 & 0.8823 \\
\hline
\multirow[c]{3}{*}{\STAB{\rotatebox[origin=c]{90}{\textsc{Stage 2a}}}} & 1-shot & 43.0 & 0.8765 & 55.0 & 0.9073 & 73.1 & 0.9382 & 57.0 & 0.9073 \\
 & 2-shot & 43.5 & 0.8785 & 54.4 & 0.9072 & 71.6 & 0.9392 & 56.5 & 0.9083 \\
 & 5-shot & 42.3 & 0.8662 & 54.4 & 0.9066 & 73.4 & 0.9347 & 56.7 & 0.9025 \\
\hline
\multirow[c]{3}{*}{\STAB{\rotatebox[origin=c]{90}{\textsc{Stage 2b}}}} & 1-shot & 44.5 & 0.8766 & 54.9 & 0.9046 & 73.1 & 0.9391 & 57.5 & 0.9068 \\
 & 2-shot & 44.5 & 0.8777 & 55.4 & 0.9080 & 74.3 & 0.939 & 58.1 & 0.9082 \\
 & 5-shot & 44.7 & 0.8734 & 55.0 & 0.9072 & 70.0 & 0.9363 & 56.6 & 0.9056 \\
\hline
\multirow[c]{3}{*}{\STAB{\rotatebox[origin=c]{90}{\textsc{Stage 3}}}} & 1-shot & \textbf{48.8} & 0.8896 & \textbf{60.5} & 0.9141 & 78.9 & \textbf{0.9480} & 62.7 & \textbf{0.9172} \\
 & 2-shot & 48.5 & \textbf{0.8914} & 60.1 & 0.9132 & \textbf{80.7} & 0.9456 & \textbf{63.1} & 0.9167 \\
 & 5-shot & 47.6 & 0.8837 & 59.0 & 0.9095 & 80.2 & 0.9437 & 62.3 & 0.9123 \\
\hline
\multirow[c]{3}{*}{\STAB{\rotatebox[origin=c]{90}{Falcon}}} &  1-shot & 31.8 & 0.8588 & 40.0 & 0.8677 & 71.6 & 0.9380 & 47.8 & 0.8882 \\
 &  2-shot & 34.5 & 0.8671 & 44.8 & 0.8876 & 76.9 & 0.9416 & 52.1 & 0.8988 \\
 &  5-shot & 40.8 & 0.8789 & X & X & 78.5 & 0.9434 & X & X \\
\end{tabular}

%% file: results_table_mdns.tex
\begin{tabular}{l|l|ll|ll|ll|ll|ll||ll}
 & & \multicolumn{2}{c}{IT} & \multicolumn{2}{c}{Koran} & \multicolumn{2}{c}{Law} & \multicolumn{2}{c}{Medical} & \multicolumn{2}{c}{Subtitles} & \multicolumn{2}{c}{Average} \\
 & & BLEU & COMET & BLEU & COMET & BLEU & COMET & BLEU & COMET & BLEU & COMET & BLEU & COMET \\
\hline
 & Baseline & 34.3 & 0.8153 & 14.7 & 0.7229 & 44.7 & 0.8696 & 43.5 & 0.8406 & 27.7 & \textbf{0.7891} & 33.0 & 0.8075 \\
\hline
\multirow[c]{3}{*}{\STAB{\rotatebox[origin=c]{90}{\textsc{Stage 0}}}} & 1-shot & 35.9 & 0.7698 & 17.2 & 0.6580 & 51.6 & 0.853 & 42.3 & 0.7964 & 17.5 & 0.6358 & 32.9 & 0.7426 \\
 & 2-shot & 35.9 & 0.7433 & 17.2 & 0.6346 & 49.9 & 0.8467 & 38.2 & 0.7810 & 22.4 & 0.7024 & 32.7 & 0.7416 \\
 & 5-shot & 31.9 & 0.7196 & 14.5 & 0.6000 & 42.3 & 0.8287 & 30.5 & 0.7505 & 24.4 & 0.7400 & 28.7 & 0.7278 \\
\hline
\multirow[c]{4}{*}{\STAB{\rotatebox[origin=c]{90}{\textsc{Stage 1}}}} & 0-shot & 39.6 & 0.8403 & 22.6 & 0.7274 & 50.7 & 0.8824 & 47.8 & 0.8429 & \textbf{28.1} & 0.7879 & 37.8 & 0.8162 \\
 & 1-shot & 36.7 & 0.7620 & 21.1 & 0.6434 & 51.1 & 0.8228 & 7.1 & 0.5078 & 0.0 & 0.4306 & 23.2 & 0.6333 \\
 & 2-shot & 35.6 & 0.7436 & 20.5 & 0.6152 & 48.9 & 0.8019 & 15.9 & 0.5441 & 0.0 & 0.4208 & 24.2 & 0.6251 \\
 & 5-shot & 32.8 & 0.7296 & 18.4 & 0.5980 & 44.9 & 0.7940 & 23.4 & 0.5854 & 16.8 & 0.6388 & 27.3 & 0.6692 \\
\hline
\multirow[c]{3}{*}{\STAB{\rotatebox[origin=c]{90}{\textsc{Stage 2a}}}} & 1-shot & 34.3 & 0.8277 & 15.5 & 0.7222 & 49.5 & 0.8739 & 43.6 & 0.8380 & 25.7 & 0.7838 & 33.7 & 0.8091 \\
 & 2-shot & 35.8 & 0.8244 & 16.4 & 0.7154 & 49.6 & 0.8739 & 44.6 & 0.8362 & 24.1 & 0.7810 & 34.1 & 0.8062 \\
 & 5-shot & 34.3 & 0.8203 & 15.9 & 0.7083 & 48.1 & 0.8659 & 40.7 & 0.8220 & 24.0 & 0.7712 & 32.6 & 0.7975 \\
\hline
\multirow[c]{3}{*}{\STAB{\rotatebox[origin=c]{90}{\textsc{Stage 2b}}}} & 1-shot & 34.6 & 0.8269 & 16.0 & 0.7217 & 50.4 & 0.8752 & 44.2 & 0.8405 & 25.9 & 0.7830 & 34.2 & 0.8095 \\
 & 2-shot & 35.5 & 0.8182 & 16.5 & 0.7150 & 49.9 & 0.8747 & 43.4 & 0.8349 & 24.5 & 0.7774 & 34.0 & 0.8040 \\
 & 5-shot & 33.5 & 0.8103 & 16.6 & 0.7070 & 48.2 & 0.8696 & 37.5 & 0.8274 & 25.2 & 0.7782 & 32.2 & 0.7985 \\
\hline
\multirow[c]{3}{*}{\STAB{\rotatebox[origin=c]{90}{\textsc{Stage 3}}}} & 1-shot & 41.4 & \textbf{0.8423} & 28.8 & 0.7235 & \textbf{58.1} & \textbf{0.8862} & \textbf{52.9} & \textbf{0.8488} & 27.0 & 0.7846 & \textbf{41.6} & \textbf{0.8171} \\
 & 2-shot & \textbf{41.7} & 0.8401 & \textbf{29.6} & 0.7225 & 57.3 & 0.8850 & 51.2 & 0.8480 & 27.6 & 0.7850 & 41.5 & 0.8161 \\
 & 5-shot & 40.9 & 0.8296 & 29.2 & 0.7249 & 55.8 & 0.8804 & 48.7 & 0.8413 & 27.5 & 0.7876 & 40.4 & 0.8128 \\
\hline
\multirow[c]{3}{*}{\STAB{\rotatebox[origin=c]{90}{Falcon}}} &  1-shot & 31.5 & 0.7985 & 17.9 & 0.7081 & 45.4 & 0.8538 & 42.4 & 0.8035 & 21.7 & 0.7586 & 31.8 & 0.7845 \\
 &  2-shot & 35.5 & 0.8202 & 22.4 & 0.7263 & 49.5 & 0.8680 & 47.5 & 0.8288 & 21.4 & 0.7605 & 35.3 & 0.8008 \\
 &  5-shot & 40.1 & 0.8377 & 24.5 & \textbf{0.7358} & 50.5 & 0.8749 & 50.1 & 0.8401 & 22.6 & 0.7776 & 37.6 & 0.8132 \\
\end{tabular}